\documentclass{article}

\usepackage{microtype}
\usepackage{graphicx}
\usepackage{subfigure}
\usepackage{booktabs} 

\renewcommand{\mid}{\mathbin|} 

\usepackage{hyperref}

\usepackage{mathrsfs}
\usepackage{algorithm}
\usepackage{caption}

\usepackage[accepted]{icml2023-diffxyz}

\usepackage{amsmath}
\usepackage{amssymb}
\usepackage{mathtools}
\usepackage{amsthm}

\usepackage[capitalize,noabbrev]{cleveref}

\usepackage{graphicx}

\theoremstyle{plain}
\newtheorem{theorem}{Theorem}[section]

\theoremstyle{definition}

\theoremstyle{remark}

\def\EE{\mathbb{E}}

\usepackage[textsize=tiny]{todonotes}

\icmltitlerunning{Differentiating Metropolis-Hastings to Optimize Intractable Densities}

\begin{document}

\twocolumn[
\icmltitle{Differentiating Metropolis-Hastings to Optimize Intractable Densities}



\icmlsetsymbol{equal}{*}

\begin{icmlauthorlist}
\icmlauthor{Gaurav Arya}{equal,mit}
\icmlauthor{Ruben Seyer}{equal,chalmers,gothenburg}
\icmlauthor{Frank Sch\"{a}fer}{mit}
\icmlauthor{Kartik Chandra}{mit}
\icmlauthor{Alexander K. Lew}{mit}
\icmlauthor{Mathieu Huot}{oxford}
\icmlauthor{Vikash K. Mansinghka}{mit}
\icmlauthor{Jonathan Ragan-Kelley}{mit}
\icmlauthor{Christopher Rackauckas}{mit,jh,pumas}
\icmlauthor{Moritz Schauer}{chalmers,gothenburg}
\end{icmlauthorlist}

\icmlaffiliation{mit}{Massachusetts Institute of Technology, USA}
\icmlaffiliation{chalmers}{Chalmers University of Technology, Sweden}
\icmlaffiliation{gothenburg}{University of Gothenburg, Sweden}
\icmlaffiliation{jh}{JuliaHub, USA}
\icmlaffiliation{pumas}{Pumas-AI, USA}
\icmlaffiliation{oxford}{University of Oxford, UK}

\icmlcorrespondingauthor{Gaurav Arya}{aryag@mit.edu}

\icmlkeywords{gradient estimation, stochastic derivatives, coupling, Monte Carlo, Metropolis-Hastings}

\vskip 0.3in
]



\printAffiliationsAndNotice{\icmlEqualContribution} 

\begin{abstract}
    We develop an algorithm for automatic differentiation of Metropolis-Hastings samplers, 
    allowing us to differentiate through probabilistic inference, even if the model has discrete components within it.
    Our approach fuses recent advances in stochastic automatic differentiation with traditional Markov chain coupling schemes, providing an unbiased and low-variance gradient estimator.
    This allows us to apply gradient-based optimization to objectives expressed as expectations over intractable target densities. We demonstrate our approach by finding an ambiguous observation in a Gaussian mixture model and by maximizing the specific heat in an Ising model.
\end{abstract}

\newif\ifintro
\newif\ifexamples
\newif\iftheory
\introtrue
\examplestrue
\theorytrue

\vspace{-1.5em}
\section{Introduction}

    Metropolis-Hastings (MH) samplers have found wide applicability across a number of disciplines due to their ability to sample from distributions with intractable normalizing constants. However, MH samplers are not traditionally differentiable due to the presence of discrete 
    accept/reject steps for proposed samples~\cite{zhang2021differentiable}. This poses a problem when we wish to optimize objectives that are themselves
    a function of the sampler's output. 
    
    In this work, we propose an approach for unbiasedly differentiating a MH sampler.
     Specifically, consider a family of distributions $\mu_\theta$ dependent on a parameter $\theta$, targeted
    by a MH sampler $\{Z_t^\theta\}_{t=1}^\infty$. 
    The samples can be used to approximate expectations of functions $f$ with respect to $\mu_\theta$; that is, if the sampler is ergodic~\cite{maruyama1959ergodic},
    \begin{align}
        &\mathop{\EE}_{X^\theta \sim \mu_\theta} \left[ f\left(X^\theta\right)\right] = 
        \lim_{T \to \infty} \frac{1}{T} \sum_{t=1}^{T} f\left(Z_t^\theta\right),
        \label{eq:ergodic}
    \end{align}
    for bounded and measurable $f$. Now, consider optimizing some function of $\mu_\theta$. For instance, \citet{chandra2022designing} consider Bayesian inference on probabilistic models of human cognition, seeking an observation $\theta$ which maximizes the variance of the posterior $\mathrm{P}(x\mid{} \theta)$ of a latent $x$.
    In such a case, we may be interested in estimating the gradient of an expectation over the density:
    \begin{equation}
        \frac{\partial}{\partial \theta} \mathop{\EE}_{X^\theta \sim \mu_\theta} \left[ f\left(X^\theta\right)\right].
        \label{eq:grad}
    \end{equation}
    This is challenging when $\mu_\theta$ 
    can only be sampled by Monte Carlo methods. One approach to estimating \eqref{eq:grad} is to differentiate through the sampling process itself. 
    However, direct application of gradient estimation strategies such as the score function estimator to MH leads to high variance as the chain length increases~\cite{thin2021monte,doucet2023differentiable}. Our key insight is that we can form a consistent estimator as the MH chain length increases, by \emph{coupling} two MH chains with perturbed targets.
    \looseness=-1

     Prior work has considered sampling procedures that avoid MH accept/reject steps~\citep{zhang2021differentiable,doucet2022score,doucet2023differentiable} or differentiated only through the continuous dynamics of samplers such as Hamiltonian Monte Carlo~\cite{chandra2022designing, campbell2021gradient, zoltowski2021slice}. 
    The latter approach leads to biased gradients
    and does not apply to discrete target distributions.
    In this work, we instead unbiasedly differentiate the accept/reject steps.
    We make the following contributions:
    \vspace{-0.5em} 
    \begin{itemize}
        \item A provably unbiased algorithm for differentiating through MH with $\mathcal{O}(1)$ multiplicative computational overhead, 
        that
        applies to arbitrary discrete or continuous target distributions, 
        based on smoothed perturbation analysis~\cite{fu1997conditional} and stochastic automatic differentiation~\cite{arya2022automatic}.
        \vspace{-0.25em}
        \item A demonstration of how Monte Carlo coupling schemes~\cite{wang2021maximal,propp1996exact} may be applied to 
        produce an efficient low-variance single-chain MH gradient estimator. 
        \vspace{-0.25em}
        \item Empirical verifications of the correctness of our gradient estimator and preliminary applications to optimizing the posterior of a Gaussian mixture model and the specific heat of an Ising model.  
    \end{itemize}

\section{Differentiable Metropolis-Hastings}

Consider the use of MH to sample from a target distribution $\mu_\theta$ using 
an unnormalized density $g_\theta(x) \propto \mu_\theta(x)$ and
a proposal density $q(x' \mid{} x)$.
At state $x = Z_t^\theta$, we draw a candidate sample $x' \sim q(\cdot \mid{} x)$ and accept it with probability 
\begin{equation}
    \alpha_\theta (x' \mid{} x) = \min\left(1, \frac{g_\theta(x')}{g_\theta(x)} \frac{q(x \mid{} x')}{q(x' \mid{} x)}\right).
\end{equation}

\cref{alg:mh} shows the corresponding algorithm\footnote{non-essential details such as burn-in are omitted.}.

\begin{algorithm}
    \begin{algorithmic}[1]
        \STATE {\bfseries Input:} functions $g_\theta, f$, proposal $q$, start state $x_1$
        \STATE $S := x_1$
        \FOR {$i=1$ {\bfseries to} $T-1$}
        \STATE {\bfseries sample} $x' \sim q(\cdot \mid{} x_i)$, $U \sim \operatorname{Unif}()$
        \STATE $b := U \le \alpha_\theta(x'\mid x_i)$
        \STATE {\bfseries if} $b = 1$ {\bfseries then} $x_{i+1} := x'$ {\bfseries else} $x_{i+1} := x_i$ {\bfseries end if}
        \STATE $S := S + f(x_{i+1})$
        \ENDFOR
        \STATE {\bfseries return} $S/T$
    \end{algorithmic}
    \caption{$T$-sample Metropolis-Hastings}
    \label{alg:mh}
\end{algorithm}

Now, let us understand the sensitivity of MH with respect to a parameter $\theta \in \mathbb R$.
The acceptance probability $\alpha_\theta(x' \mid{} x)$ depends on $\theta$, with derivative in the non-trivial case
\begin{equation}
    \frac{\partial}{\partial \theta} \alpha_\theta (x' \mid{} x)= \alpha_\theta (x' \mid{} x)\frac{\partial}{\partial \theta} \log \frac{g_\theta(x')}{g_\theta(x)}.
\end{equation}
This acceptance probability feeds into the discrete random accept/reject step (\cref{alg:mh}, lines 4-5).
A number of gradient estimation approaches have been developed in such a discrete random setting, including
score-function estimators~\cite{kleijnen1996optimization}, measure-valued derivatives~\cite{heidergott2008measure}, and smoothed perturbation analysis (SPA)~\cite{fu1997conditional}.
In this work, we opt for an SPA-based estimator, which for a purely discrete random variable $X^\theta$ assumes the following form~\cite{heidergott2008measure,arya2022automatic}:
\looseness=-1
\begin{equation}
   \frac{\partial}{\partial \theta} \EE[f(X^\theta)]= \EE\left[w^\theta \left(f(Y^\theta) - f(X^\theta)\right)\right].
   \label{eq:est}
\end{equation}
Intuitively, the estimator works by taking differences between the program evaluated at the primal sample $X^\theta$ and at a discretely perturbed alternative sample $Y^\theta$ that ``branches off'' the primal, weighting these by a possibly random \(w^\theta\) related to the infinitesimal probability of a change.
In the case of the long Markov chains produced by MH,
we will see that the coupling of $X^\theta$ and $Y^\theta$, i.e. the form of their joint distribution, plays a key role in reducing variance.

There has been recent interest in automating the application of such strategies across whole programs~\cite{lew2023adev,arya2022automatic,krieken2021storchastic}.
In particular, \citet{arya2022automatic} develop composition rules for an SPA-based estimator to perform automatic unbiased gradient estimation for discrete random programs, calling their
construction ``stochastic derivatives.''
We use it to develop a differentiable MH sampler, given in \cref{alg:dmh}.
\looseness=-1

\definecolor{comment}{rgb}{0.0, 0.46, 0.37}
\begin{algorithm}
    \caption{$T$-sample Differentiable Metropolis-Hastings}
    \begin{algorithmic}[1]
        \STATE {\bfseries Input:} functions $g_\theta, f$, proposal $q$, start state $x_1$,
        \STATE \hspace{2.85em} coupled proposal $q_{xy}$
        \STATE $\partial S := 0$, $y_1 := x_1$, $w_1 := 0$ 
        \FOR {$i=1$ {\bfseries to} $T-1$}
        \STATE {\color{comment} // Perform MH step for primal and alternative}
        \STATE {\bfseries sample} $x', y' \sim q_{xy}(\cdot, \cdot \mid{} x_i, y_i)$, $U \sim \operatorname{Unif}()$
        \STATE $b_x := U \le \alpha_\theta(x'\mid x_i)$, $b_y := U \le \alpha_\theta(y'\mid y_i)$, 
        \STATE {\bfseries if} $b_x = 1$ {\bfseries then} $x_{i+1} := x'$ {\bfseries else} $x_{i+1} = x_i$ {\bfseries end if}
        \STATE {\bfseries if} $b_y = 1$ {\bfseries then} $y_{i+1} := y'$ {\bfseries else} $y_{i+1} := y_i$ {\bfseries end if}
        \STATE {\color{comment} // Compute stochastic derivative weight}
        \IF {$b_x = 1$}
            \STATE $w := {\alpha_\theta(x' \mid x_i)}^{-1} \max(0, -\frac{\partial \alpha_\theta(x' \mid x_i)}{\partial \theta})$ 
        \ELSE
            \STATE $w := {(1-\alpha_\theta(x' \mid x_i))}^{-1} \max(0, \frac{\partial \alpha_\theta(x' \mid x_i)}{\partial \theta})$ 
        \ENDIF
        \STATE {\color{comment} // Prune between previous alternative and new}
        \STATE {\bfseries if} $y_{i+1} = x_{i+1}$ {\bfseries then} $w_i := 0$ {\bfseries end if} {\color{comment}// drop recoupled}
        \STATE $w_{i+1} := w + w_i$
        \STATE {\bfseries sample} $\varpi \sim \operatorname{Unif}()$
        \IF {$\varpi \cdot w_{i+1} < w$}
            \STATE {\bfseries if} $b_x = 1$ {\bfseries then} $y_{i+1} := x_i$ {\bfseries else} $y_{i+1} = x'$ {\bfseries end if}
        \ENDIF
        \STATE {\color{comment} // Update derivative estimate}
        \STATE $\partial S := \partial S + w_{i+1}(f(y_{i+1}) - f(x_{i+1}))$
        \ENDFOR
        \STATE {\bfseries return} $\partial S/T$
    \end{algorithmic}
    \label{alg:dmh}
\end{algorithm}

Most parts of \cref{alg:dmh} follow from directly applying the composition rules of \citet{arya2022automatic} to \cref{alg:mh}.
At a high level, alternative MH samples $y_i$ are propagated in parallel to the primal samples $x_i$ (lines 6--9). 
For each primal accept/reject step $b_x$, we compute the weight of a flip in $b_x$ according to the stochastic derivative estimator (lines 11--15).
We employ the pruning strategy given in \citet{arya2022automatic} to stochastically select a single alternative between the currently tracked alternative and the new possible alternative (lines 17--22), with the extra optimization that we stop tracking alternatives that have recoupled, as they will stay coupled for all future steps and no longer contribute to the derivative.
On recoupling, we can thus always prune and consider a new alternative.
Note that \cref{alg:dmh} accepts a ``coupled proposal'' $q_{xy}$, which specifies the joint proposal distribution for the primal and alternative MH chains (line 6). As long as $q_{xy}$ is a valid coupling, \cref{alg:dmh} computes an unbiased derivative estimate of finite-sample MH expectations:
\begin{theorem}\label{thm:unbiased}
    Suppose that for all $x,y$, it holds that if $x', y' \sim q_{xy}(\cdot, \cdot \mid{} x, y)$ then $x' \sim q(\cdot \mid{} x)$ and $y' \sim q(\cdot \mid{} y)$ (i.e.~$q_{xy}$ is a proposal coupling),
    and furthermore that if $x = y$ then $x' = y'$.
    Then, with inputs $g_\theta(x) \propto \mu_\theta(x)$, a bounded and measurable $f$, $q$, $x_0$, and $q_{xy}$, 
    the output of \cref{alg:dmh} is an unbiased estimator of\begin{equation}
    \frac{\partial}{\partial \theta} \left(\EE\left[\frac{1}{T} \sum_{i=1}^{T} f\left(Z_t^\theta\right)\right]\right),
    \end{equation}\\
    where $Z_t^\theta$ is a MH sampler of $\mu_\theta$ with proposal $q$, initial state $Z_0^\theta = x_0$.
    \label{thm:thm}
\end{theorem}

\begin{figure}
    \centering
    \includegraphics[trim={0 2cm 0 0},clip,width=\linewidth]{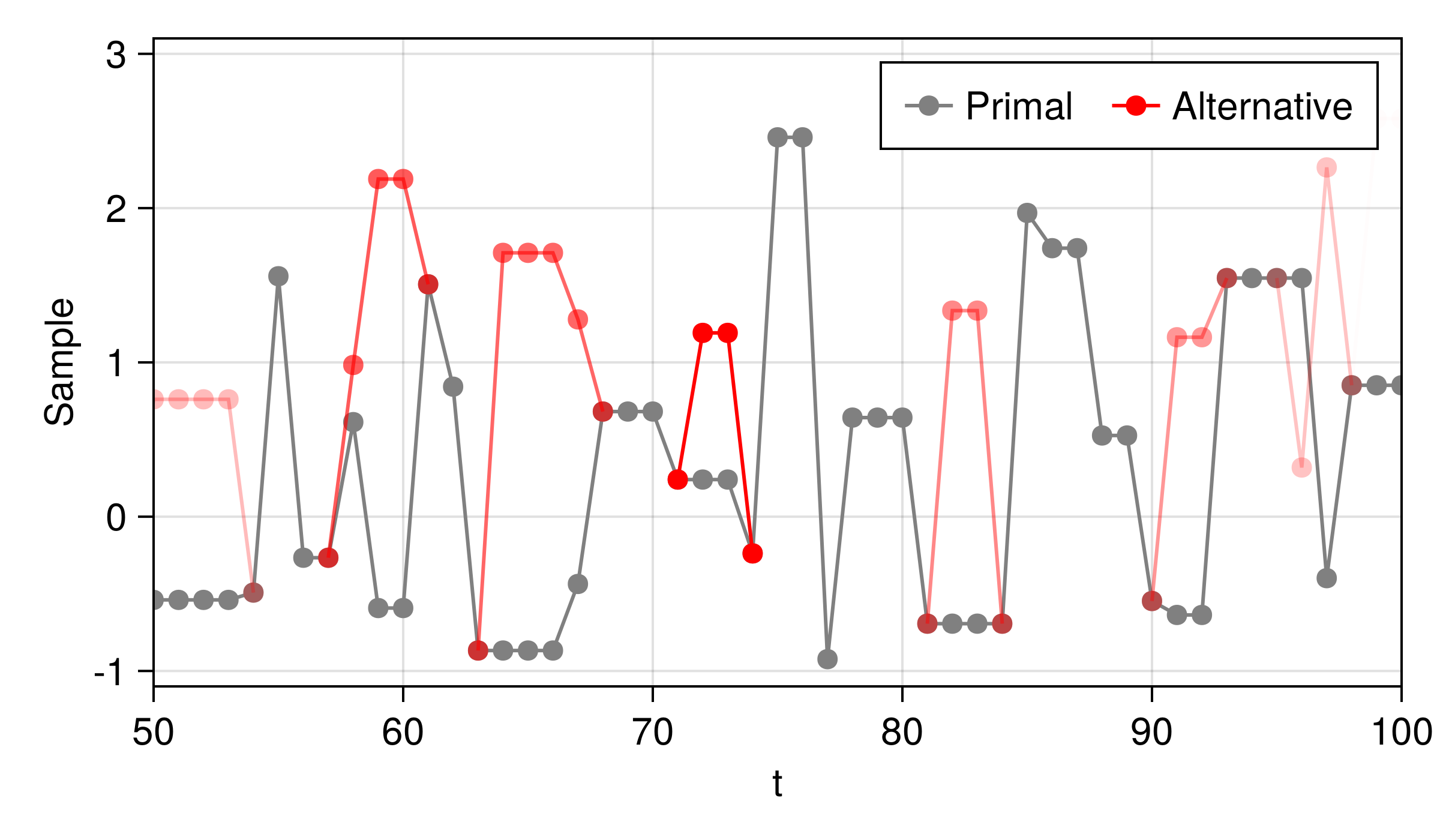}
    \vspace{-1.5em}
    \caption{Differentiable MH (\cref{alg:dmh}) samples two coupled chains: the primal chain (grey) and a chain containing alternative samples (red), which together allow to estimate the derivative of the sampler. Here 
    we target $\mu_\theta \sim \mathcal{N}(\theta, 1)$ with $\theta = 0.5$ and a maximal reflection proposal coupling.
    The alpha values of the depicted alternative paths correspond to their weight. }
    \label{fig:perturbations} 
\end{figure}

While unbiasedness is guaranteed by \cref{thm:thm}, the choice of coupling is important for the variance performance of \cref{alg:dmh}.
A simple choice is \emph{common random numbers} (CRN) \cite{glasserman1992guidelines}, equivalent to using the same random seed for both chains: we use CRN for the accept/reject step in \cref{alg:dmh}.
For the proposal, we leverage prior work on coupling for its more traditional use: minimizing recoupling time. That is, if alternative trajectories rapidly recouple to the primal, we will be able to consider additional alternative trajectories over the lifetime of the chain, hence reducing variance.
\citet{wang2021maximal} present several schemes for coupling MH proposals. \cref{fig:perturbations} gives an example of the \emph{maximal reflection coupling} for a Gaussian proposal applied in \cref{alg:dmh}: we see that the alternative trajectories recouple within $\approx 5$ steps. 
\looseness=-1

Ultimately, we note that \cref{alg:dmh} can be derived automatically from \cref{alg:mh} and need not be handwritten;
in code, we can implement our differentiable MH by applying
\texttt{StochasticAD.jl} of \citet{arya2022automatic} 
to \cref{alg:mh}.
Only the proposal coupling needs to be manually specified, if it differs from CRN.

\section{Examples}
\label{sec:examples}

\ifexamples
\subsection{Finding ambiguous observations in a Gaussian mixture model}
\label{sec:mixture}

\begin{figure}
    \centering
    \includegraphics[trim={0 2cm 0 0},clip,width=\linewidth]{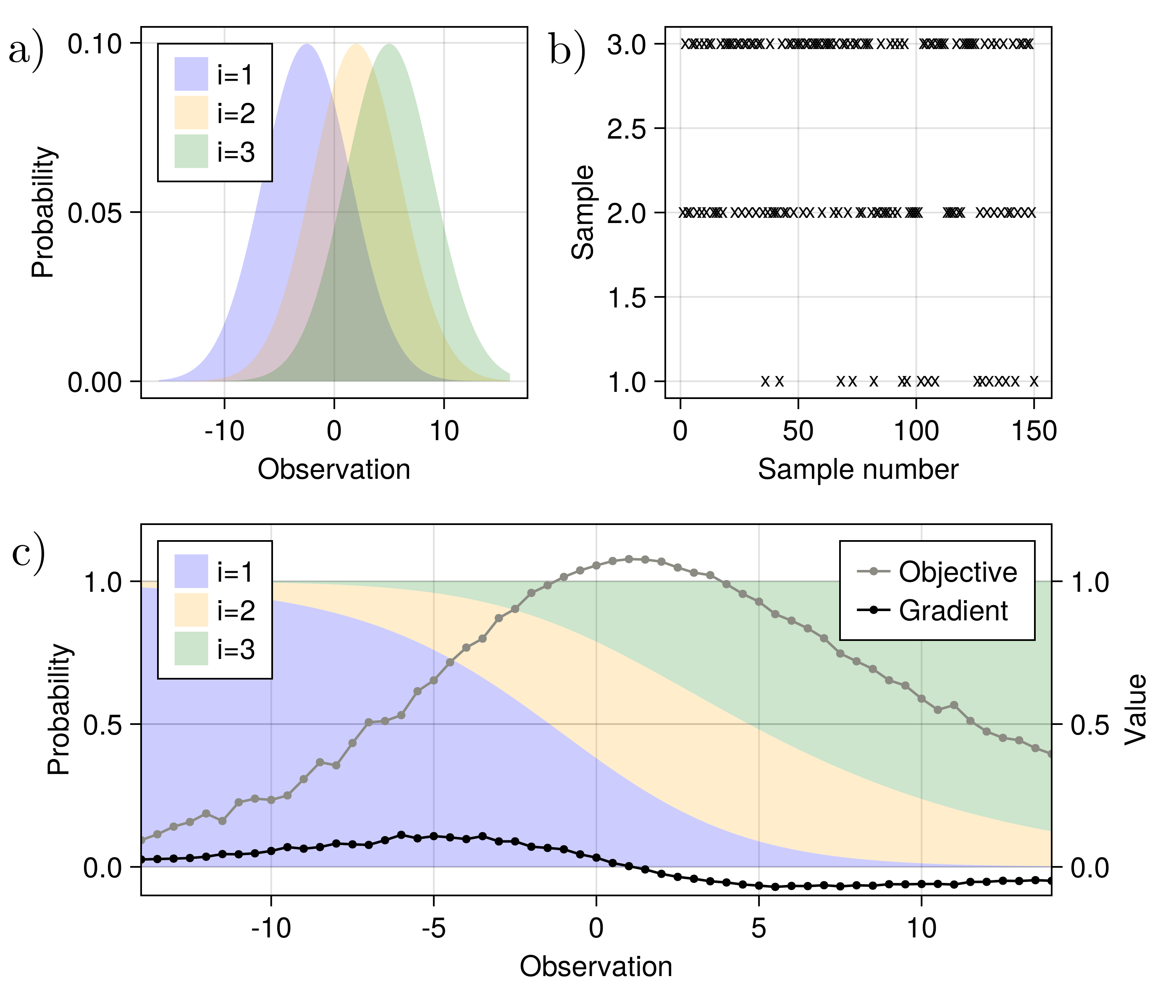}
    \vspace{-1.5em}
    \caption{Finding ambiguous observations in a Gaussian
mixture model by differentiating MH inference.
     a) Density of each Gaussian component.
     b) Samples from an MH chain inferring the component $J$ for the observation $H = 4.0$.
     c) Posterior density of $J$ for each observation. The values of the optimization objective (posterior entropy) and its estimated gradient are overlayed. 
     }
    \label{fig:mcmc_mixtures} 
\end{figure}

Consider a mixture model with three independent Gaussians $N_i \sim N(\mu_i, \sigma)$ with means $\mu_1 = -2.5, \mu_2 = 2, \mu_3 = 5$, and $\sigma = 4$ [\cref{fig:mcmc_mixtures}a)]. Suppose we pick a component $J \in \{1,2,3\}$ uniformly at random and sample $H \sim N_J$.
Now, conditional on the observation $H$, e.g.~the height of a person in a population,
we would like to infer the source cluster $J$. 
By Bayes' rule,  the posterior [\cref{fig:mcmc_mixtures}c)] is
\begin{equation}
    \mathrm{P}(J = j \mid{} H = h) \propto \mathrm{P}(H = h\mid{} J = j) \mathrm{P}(J = j).
    \label{eq:prop}
\end{equation}
In this case, the posterior has a small finite support, allowing us to compute 
the usually intractable normalization constant of \eqref{eq:prop} via explicit enumeration.
However, in order to test our approach, we consider sampling the posterior by MH (\cref{fig:mcmc_mixtures}b)
using only the unnormalized density (\ref{eq:prop}). 

Now, we use differentiable MH to optimize the posterior distribution. We seek to find the observation $h$ that
maximizes inference ambiguity, i.e. finds the observation for which it is most difficult to determine which cluster it came from by maximizing posterior entropy.
Even in this toy setting, a na\"ive application of the score function estimator yields an estimator whose variance does not decrease with more MH samples (\cref{sec:score}).
However, \cref{alg:dmh}, using \emph{maximal independent} proposal coupling~\cite{vaserstein1969markov}, does better. This coupling seeks to maximize the probability that the proposals agree and recouple on the next step. A sweep of the objective and its gradient, computed by differentiable MH, is depicted in \cref{fig:mcmc_mixtures}c. Optimization via gradient descent converges to the 
optimum observation in $\approx 100$ iterations. 
\looseness=-1

\subsection{Maximizing the specific heat in an Ising model}

\begin{figure}
    \centering
    \includegraphics[width=\linewidth]{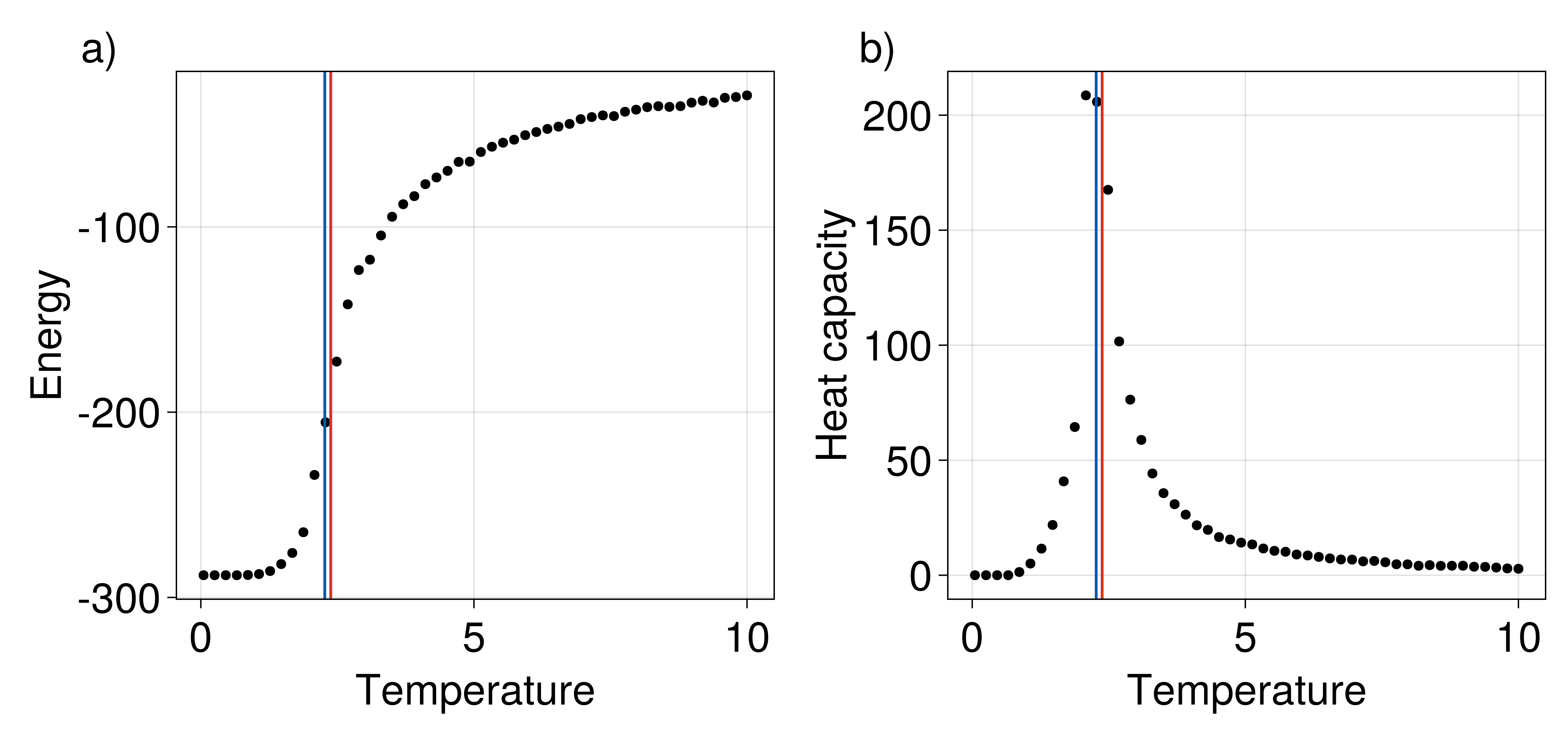}
    \vspace{-1.5em}
    \caption{Maximizing the specific heat in an Ising model as a function of temperature $T$ via differentiable MH, for $L=12$ and $\theta=1$. a) Average energy $\mathop{\EE}
    \left[ H(X^T, \theta)\right]$. b) Heat capacity \eqref{eq:specific_heat}. The blue and red lines correspond to the analytical value~\eqref{eq:Ising_analytical} and the numerically optimized prediction of the critical temperature, respectively.}
\label{fig:mcmc_ising} 
\end{figure}

Next, we consider an example from physics. The random configuration $X^T$ of a classical physical system at thermal equilibrium in contact with a large thermal reservoir of temperature $T$,  follows the Boltzmann distribution,  
\begin{equation}
   \mathrm{P}(X^T = x\mid{} \theta) = e^{-{H}({x},\theta)/(k_B T)}/Z(\theta),
\end{equation}
where $Z(\theta)
 = \sum_x e^{-{H}({x},\theta)/(k_B T)}
$ is the partition function.
Consider the two-dimensional isotropic Ising model for spin configurations on an $L {\times} L$ square lattice with periodic boundary conditions. 
The spin $x_{j,k}$ at a site $(j,k)$ 
can take a value of either $+1$ or $-1$, resulting in a state space of size $2^{L^2}$. The Hamiltonian for this model is given by
\begin{equation}\textstyle
    H(x,\theta)= -\theta \sum_{j,k=1}^L  \left( x_{j,k} x_{j, k+1} + x_{j+1,k} x_{j, k}\right),
    \label{eq:Ising_model}
\end{equation}
where the parameter $\theta$ represents the strength of the nearest-neighbor interaction. The identification of phase transitions is central to understanding the properties and behavior of a wide range of material systems~\cite{arnold:2022}. The Ising model exhibits a symmetry-breaking phase transition at a critical temperature of
\begin{equation}
    T_c = {2\theta}/({k_B \log(1+\sqrt{2})}),
    \label{eq:Ising_analytical}
\end{equation}
in the limit $L\rightarrow\infty$, between an ordered (low temperature) and a disordered (high temperature) phase. This phase transition is associated with a peak in the heat capacity
\begin{align}
    C(T)  
= \mathop{\operatorname{Var}}\!\left( H(X^T, \theta)\right)/(k_B T^2).
    \label{eq:specific_heat}
\end{align}

\begin{figure}
    \centering
    \includegraphics[width=0.6\linewidth]{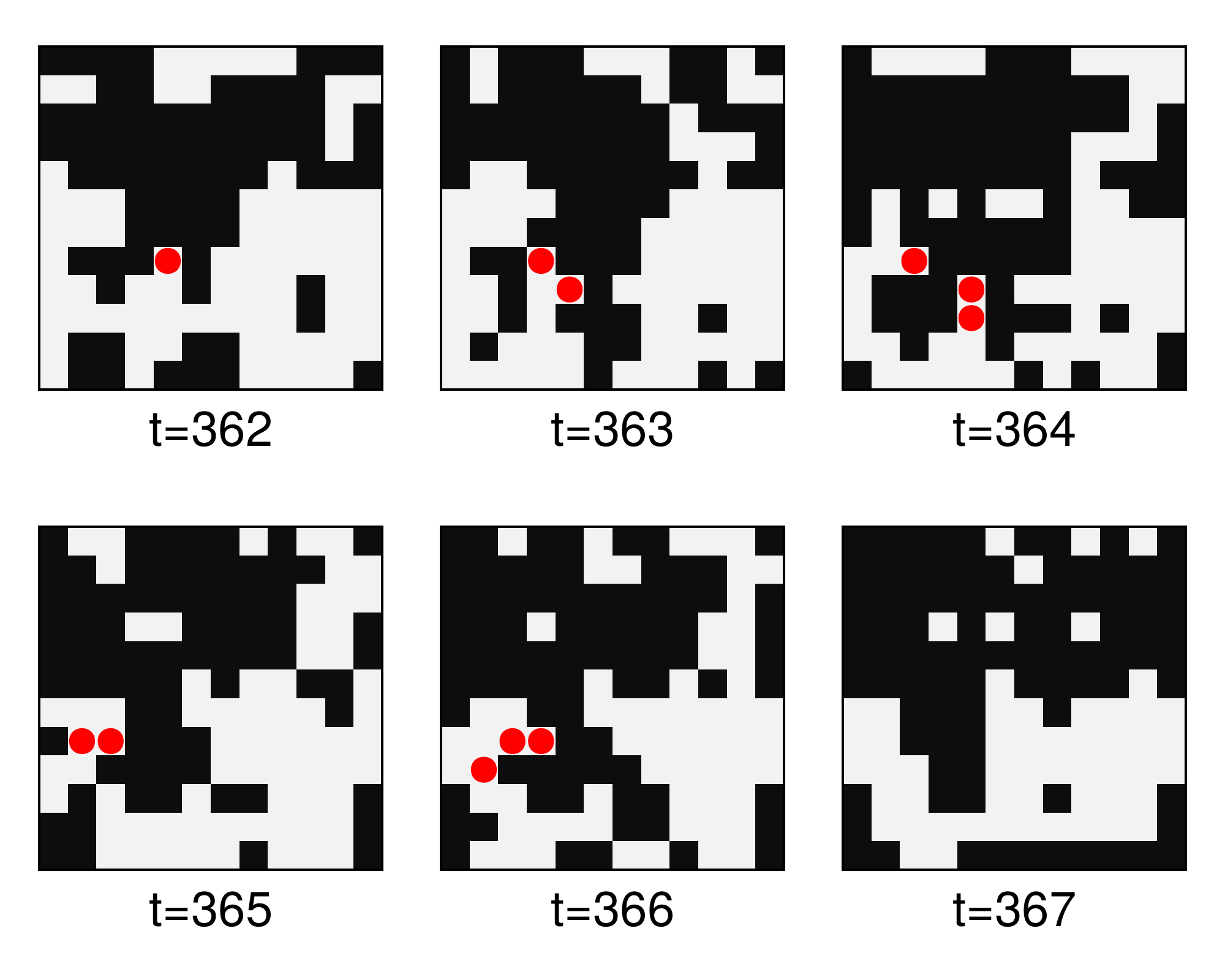}
    \vspace{-1em}
    \caption{Recoupling of primal (black and white) and alternative (flipped cells in red) spin states in the Ising model over a subrange of sweeps in the heat bath algorithm, with $L = 12$.}
\label{fig:mcmc_ising_recouple} 
\end{figure}

Our goal is to find the critical temperature by maximizing the heat capacity. 
In general, computing $C(T)$ by exhaustive enumeration is not feasible due to the size of the configuration space, but it can be computed with a Monte Carlo procedure. 
To sample configurations we use a variant of the heat bath algorithm in which we pick a site, propose to set the spin at that site to $+1$ or $-1$ with equal probability and use a MH step to decide whether to accept this proposal. 
We couple by proposing the same change for primal and alternative together with common random numbers to check for acceptance in both chains.
It is easy to see that this creates a monotone coupling \citep{propp1996exact}.
As the primal and alternative at the time of a branch are ordered with respect to the natural partial order on the configuration space, this order is preserved until the branches recouple after a finite time (\cref{fig:mcmc_ising_recouple}). Having a differentiable MH sampler for $C(T)$, we can then perform stochastic gradient ascent to find the optimal value of $T$, as shown in \cref{fig:mcmc_ising} (optimization trace provided in \cref{sec:optheat}).
\looseness=-1

\section{Conclusion and outlook}

We have presented an efficient low-variance derivative estimator for Metropolis-Hastings samplers 
and showed its efficacy in discrete and high-dimensional spaces. A key avenue for future work is to apply our scheme in settings with many trainable parameters, for example optimizing over models conditioned on observed images and videos \citep{chandra2022designing, chandra2023acting}, training energy-based models \citep{du2020improved}, and working with nested models \citep{zhang2022reasoning}. This may require incorporating reverse-mode automatic differentiation and exploring further variance reduction techniques. We may also wish to unbiasedly differentiate samplers with both discrete and continuous dynamics, see e.g. concurrent work on differentiating piecewise deterministic Monte Carlo samplers~\cite{seyer2023differentiable}. Additionally, for applications such as gradient-based hyperparameter optimization \citep{campbell2021gradient}, we may also want to differentiate with respect to parameters of the proposal and support alternative optimization objectives such as autocorrelations of the MH chain.

\else \fi 

\bibliography{example_paper}
\bibliographystyle{icml2023}

\onecolumn
\appendix
\section{Appendix}

\subsection{Differentiable Metropolis-Hastings via Score Method}
\label{sec:score}

\begin{figure*}[hb]
\centering
\begin{minipage}{0.48\textwidth}
\begin{algorithm}[H]
    \caption{$T$-sample Differentiable Metropolis-Hastings via Score}
    \begin{algorithmic}[1]
        \STATE {\bfseries Input:} functions $g_\theta, f$, proposal $q$, start state $x_1$
        \STATE $\partial S := 0$, $w := 0$
        \FOR {$i=1$ {\bfseries to} $T-1$}
        \STATE {\bfseries sample} $x' \sim q(\cdot \mid{} x_i)$, $U \sim \operatorname{Unif}()$
        \STATE $b := U \le \alpha_\theta(x'\mid x_i)$
        \STATE {\bfseries if} $b = 1$ {\bfseries then} $x_{i+1} := x'$ {\bfseries else} $x_{i+1} := x_i$ {\bfseries end if}
        \STATE $w := w + \dfrac{\partial \log \alpha_\theta(x_{i+1}\mid x_i)}{\partial \theta}$
        \STATE $\partial S := \partial S + w \cdot f(x_{i+1})$
        \ENDFOR
        \STATE {\bfseries return} $\partial S/T$
    \end{algorithmic}
    \label{alg:mhs}
\end{algorithm}
\end{minipage}
\hfill
\begin{minipage}{0.48\textwidth}
\begin{figure}[H]
    \centering
    \includegraphics[width=0.84\linewidth]{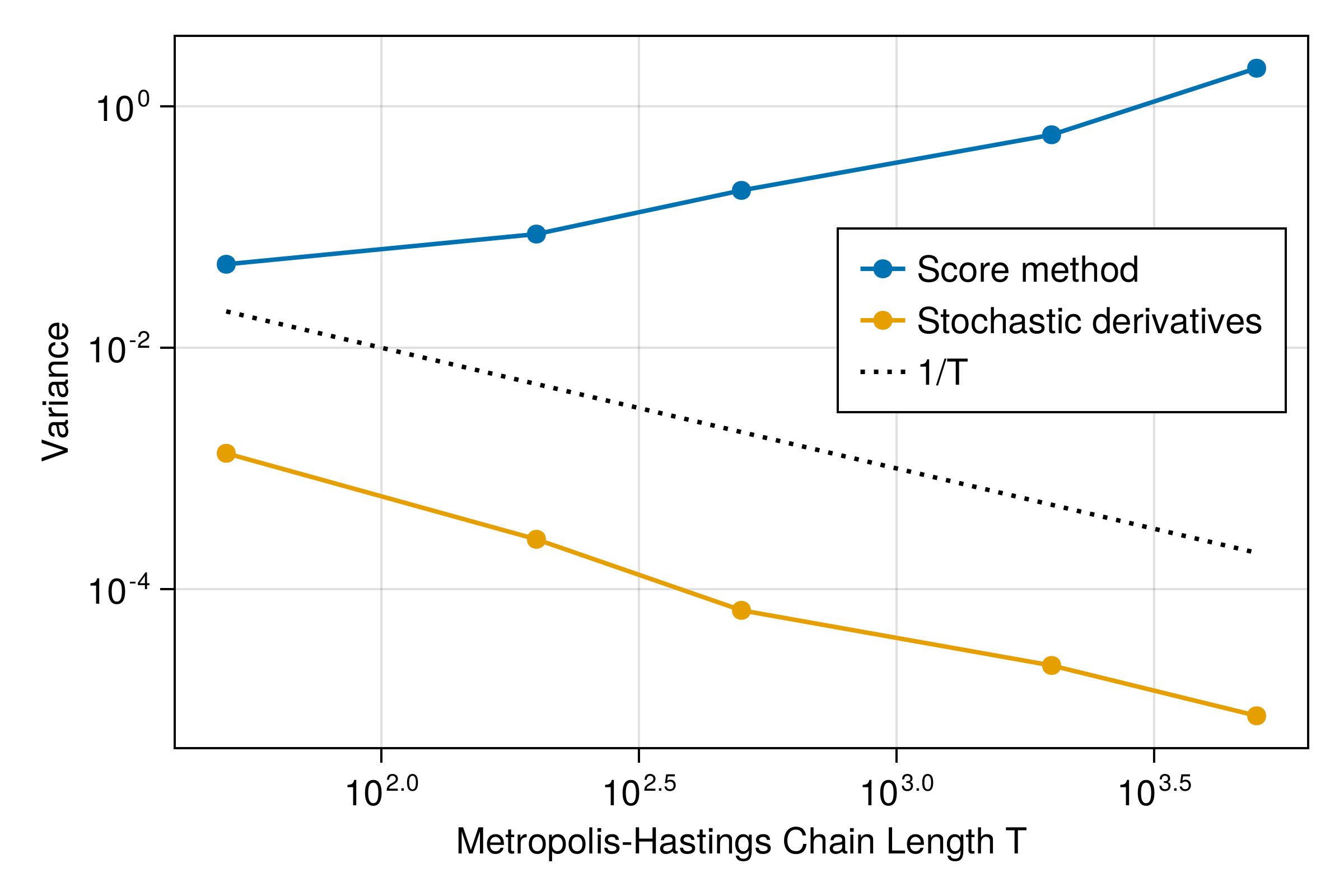}
    \vspace{-1.5em}
    \caption{Variance comparison of stochastic derivative-based differentiable MH (\cref{alg:dmh}) and score method-based differentiable MH (\cref{alg:mhs}), for the Gaussian mixture inference task given in \cref{sec:mixture} with the observation $H = 0.4$.}
\label{fig:score} 
\end{figure}
\end{minipage}
\end{figure*}
\cref{alg:mhs} applies a score-function estimator to each accept/reject step of \cref{alg:mh}.
Note that we cannot apply the score-function estimator directly to the density $\mu_\theta$, since the normalizing constant is unknown and dependent on the parameter $\theta$.
Thus, just as we did with the stochastic derivative estimator in \cref{alg:dmh}, we apply the score function estimator to each step of \cref{alg:mh} to obtain an unbiased estimator of finite-sample expectations for comparison.

\subsection{Proof of \cref{thm:unbiased}}
\begin{proof}
   Since Algorithm \ref{alg:mh} implements MH, and Algorithm \ref{alg:dmh} 
   follows from applying the composition rules of \citet{arya2022automatic} to each step of \cref{alg:mh},
   and then applying \eqref{eq:est} to compute $\partial S$,
   the result follows from Theorem 2.6 (Chain Rule) and Proposition 2.3 (Unbiasedness) of \citet{arya2022automatic}. 
   One additional trick employed by \cref{alg:dmh} is to drop the currently tracked alternative 
   if it has recoupled (line 17), which is justified by the assumption that $x = y$ implies $x' = y'$
   and the fact that for the common random numbers coupling $\alpha(y' \mid y) = \alpha(x'\mid x)$ implies $b_x = b_y$, so that chains that recouple are guaranteed to remain coupled and have no further derivative contribution in \cref{eq:est}.
\end{proof}

\subsection{Optimization of the heat capacity in the Ising model}
\label{sec:optheat}

\begin{figure}[hb]
    \centering
    \includegraphics[width=0.5\linewidth]{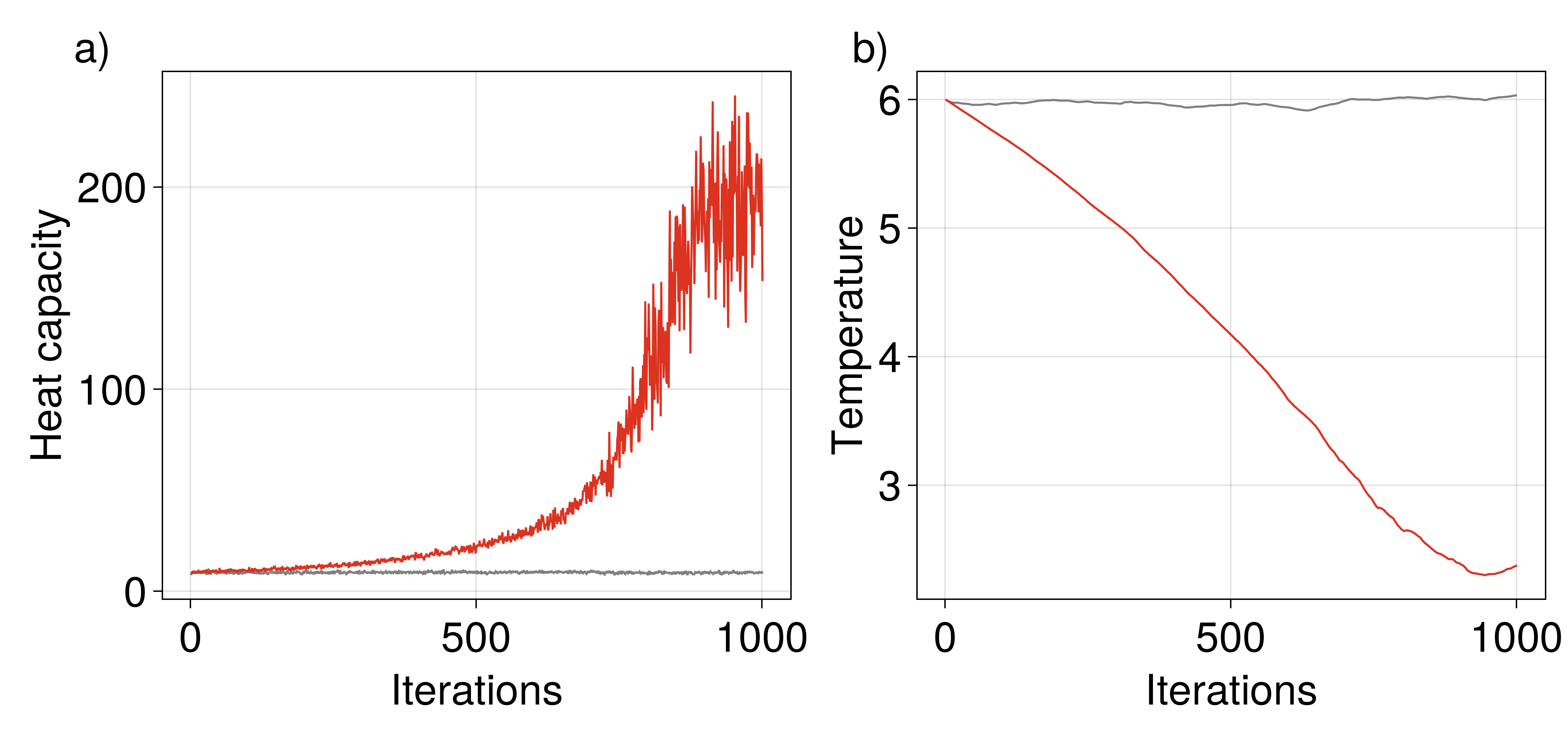}
    \vspace{-1.5em}
    \caption{Maximizing the specific heat in an Ising model~\eqref{eq:Ising_model} with respect to temperature $T$ via differentiable MH, for $L=12$ and $\theta=1$. a) Heat capacity \eqref{eq:specific_heat} across Adam optimizer iterations. b) Temperature across Adam optimizer iterations. The red and gray lines correspond to coupled and uncoupled proposals in \cref{alg:dmh}, respectively.}
\label{fig:mcmc_ising_2} 
\end{figure}

\subsection{Code}

Code for reproducing the experiments in this paper is available at \url{https://github.com/gaurav-arya/differentiable_mh}.

\end{document}